\documentclass{article}
\usepackage{ijcai25}

\usepackage{times}
\usepackage{soul}
\usepackage{url}
\usepackage[hidelinks]{hyperref}
\usepackage[utf8]{inputenc}
\usepackage[small]{caption}
\usepackage{graphicx}
\usepackage{amsmath}
\usepackage{amsthm}
\usepackage{booktabs}
\usepackage{algorithm}
\usepackage{algorithmic}
\usepackage[switch]{lineno}

\usepackage{caption}
\usepackage{subfigure}
\usepackage{pgfplots}
\usepackage{booktabs,tabularx, colortbl} 
\usepackage{amsfonts}
\usepackage{graphicx}
\usepackage{multirow}
\usepackage{enumitem}
\title{TreeKV: Smooth Key-Value Cache Compression with Tree Structures}

\author{
Ziwei He$^{1*}$,
Jian Yuan$^{1}$\thanks{Equal contribution.},
Haoli Bai$^{2}$, 
Jingwen Leng$^{1}$,
Bo Jiang$^{1}$\thanks{Corresponding author.}
\\
$^{1}$Shanghai Jiao Tong University
\quad $^{2}$Huawei Noah's Ark Lab
\\
\{ ziwei.he, yuanjian, leng-jw, bjiang\}@sjtu.edu.cn
}
\pgfplotsset{compat=1.18}  
\begin{document}

\maketitle

\begin{abstract}
Efficient key-value (KV) cache compression is critical for scaling transformer-based Large Language Models (LLMs) in long sequences and resource-limited settings. Existing methods evict tokens based on their positions or importance, but position-based strategies can miss crucial information outside predefined regions, while those relying on global importance scores resulting in strong regional biases, limiting the KV cache's overall context retention and potentially impairing the performance of LLMs on complex tasks. Our wavelet analysis reveals that as tokens approach the end of sequence, their contributions to generation gradually increase and tends to diverge more from neighboring tokens, indicating a smooth transition with increasing complexity and variability from distant to nearby context. Motivated by this observation, we propose TreeKV, an intuitive, training-free method that employs a tree structure for smooth cache compression. TreeKV maintains a fixed cache size, allowing LLMs to deliver high-quality output in long text scenarios and is applicable during both the generation and prefilling stages. TreeKV consistently surpasses all baseline models in language modeling tasks on PG19 and OpenWebText2, allowing LLMs trained with short context window to generalize to longer window with a 16x cache reduction. On the Longbench benchmark, TreeKV achieves the best performance with only 6\% of the budget at optimal efficiency\footnote{https://github.com/ZiweiHe/TreeKV}.
\end{abstract}

\section{Introduction}
\label{intro}
Large Language Models (LLMs) exhibit impressive ability to comprehend and produce text at human level, enabling them to perform tasks such as summarization, question answering, and creative writing \cite{wei2022emergent,yuan2022wordcraft,zhang2024benchmarking}. To support efficient token generation, transformer-based LLMs typically store the key-value (KV) pairs of past tokens in memory, referred to as KV cache \cite{pope2022efficiently}. However, for very long sequences,  the KV cache can require a memory that is several times larger than that for storing the model parameters, posing significant challenges in long context scenarios or resource-limited environments \cite{liu2024scissorhands}. This necessitates innovative strategies to optimize the KV cache memory footprint without compromising the performance.

Aiming for training-free methods, recent studies propose to cache the KV pairs of only a subset of past tokens, subject to a pre-defined capacity constraint.  
Those efficient KV cache methods typically address two different stages, prefilling stage and decoding stage. The prefilling stage is primarily the phase where the model develops a representation of the context it receives. The decoding stage focused on generating text output based on the encoded input. Effective management of both stages ensures that the cache operates not just as a temporary storage solution but as a proactive element enhancing overall system efficiency. 
For decoding stage optimization, some methods \cite{xiao2023efficient,han2024lm} retain only initial and recent tokens, while others \cite{zhang2024h2o,oren2024transformers,liu2024scissorhands} select tokens based on their importance scores. Those approaches help manage cache size and allow models to generate sequences longer than the pre-training context length. However, these methods often lack efficient prefilling strategy, leading to high latency due to token-by-token processing. In contrast, there are other methods focusing on optimizing the prefilling stage strategy by intelligently populating the cache based on the input context. For example, \cite{li2024snapkv} retained only critical tokens from the context, while \cite{zhang2024pyramidkv,wang2024xl3m} identified key information from KV cache using structural insights.

Despite their relative success, existing cache eviction methods have their limitations. The pure position-based selection strategies may miss out important tokens outside the pre-defined regions. On the other hand, as we will see in Figure \ref{fig:observations}, the strategies based on importance scores turn out to have very strong regional bias, which will limit the KV cache's ability to maintain a global view and potentially impair LLMs performance on complex, context-rich tasks.

This work aims to overcome the above limitations. To gain a clearer understanding of the characteristics conveyed by the KV cache, we use wavelet transform to examine the frequency representation of information contributed by tokens at different positions during generation.
The results reveal that as tokens approach the end of the sequence, the amplitudes of signals corresponding to different frequency components gradually increase, particularly at higher frequencies. This suggests that the information contributed by a token not only increases, but also becomes more distinct from its neighboring tokens as it approaches the end, indicating a smooth transition with increasing complexity and variability from distant to nearby context. This observation inspired our approach, leading to the design of a structure that is sparse on the left and dense on the right.

Based on the insights, we propose TreeKV, an intuitive, training-free approach that employs a tree structure for smooth cache compression. Unlike other cache eviction strategies, TreeKV optimizes computational efficiency and memory usage by maintaining an abstraction of the input sequence, facilitating structured and smooth transitions in context granularity between short-range and long-range contexts. By strategically removing tokens from the distant past while prioritizing the recent, proposed approach minimizes bias from heavily concentrated regions, enhancing the model's ability to handle tasks requiring comprehensive context. TreeKV distinguishes itself from most cache compression methods by being applicable to both the generation and prefilling phases, facilitating long-form generation and nuanced long context understanding.

Our contributions are summarized as follows:
\begin{itemize}
    \item We analyze the frequency representations of information collected during generation using wavelet decomposition. The results show that as tokens near the end of sequence, all frequency components gradually increase, particularly at higher frequencies. This insight inspired our method, resulting in a structure designed to be sparse on the left and dense on the right.
    \item We introduce TreeKV, an innovative, training-free method that employs a tree structure to enhance smooth cache compression. Our ablation study further proved the tree structure's significant role in shaping the model's decision making.
     \item We provide extensive experimental results that validate the effectiveness of TreeKV in both prefilling and generation stages. 
     In langauge modeling task, TreeKV allows LLMs to generalize to sequences of at least 16k, attaining the lowest perplexity among all baseline models. On the Longbench benchmark, TreeKV consistently surpasses other methods across all cache sizes, using only 6\% of budget at optimal efficiency.
\end{itemize}

\section{Related Work}\label{sec:related}
In recent years, the field of natural language processing has seen a surge in research addressing the challenges associated with KV cache eviction and memory compression in transformer-based architectures. 

StreamingLLM \cite{xiao2023efficient} and LM-Infinite \cite{han2024lm} identify that attention scores mainly concentrate on initial and recent tokens in the KV cache, leading their methods to retain only those tokens to reduce cache size. However, this could result in the loss of significant information by discarding tokens between the initial and sliding window. H2O \cite{zhang2024h2o} introduces a cache eviction method that greedily selects tokens based on importance scores derived from cumulative attention weights during generation. Scissorhands uses a similar strategy but binarizes the scores. TOVA \cite{oren2024transformers} selects tokens using the last token's attention scores. These methods, however, often overlook the structure of key information distribution by naively evicting tokens across the entire sequence.

\begin{figure}[t]
    \centering
    \includegraphics[width=0.88\linewidth]{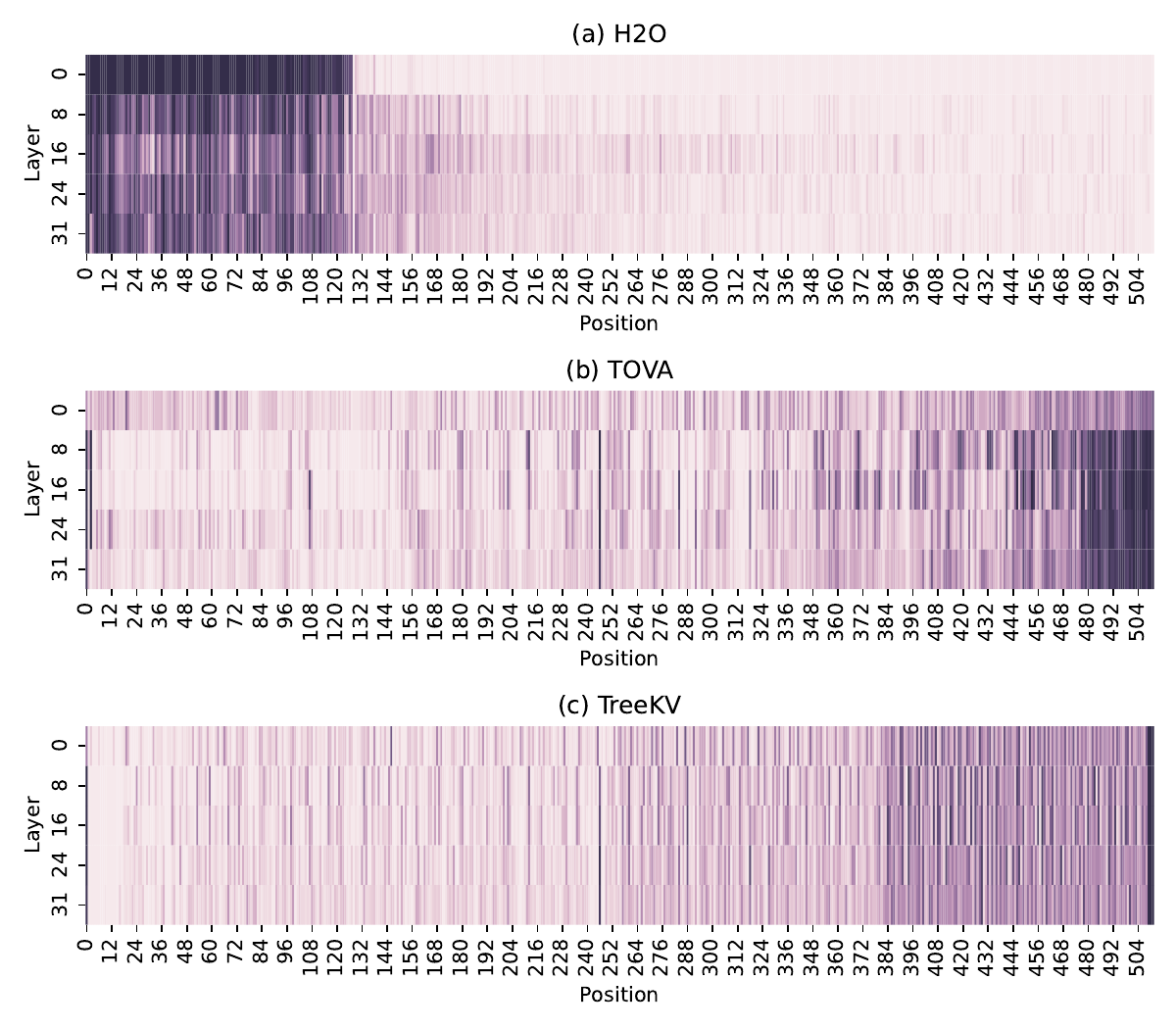}
    \caption{Distribution map of tokens selected by H2O, TOVA and TreeKV, using a cache size of 128 on a 512-length sequence from PG19. We quantify the values of selected token positions as 1 (dark color) and 0 (light color) otherwise, then average across the 32 heads to reduce noise. Note that sinks and recent tokens are not particularly kept.}
    \label{fig:observations}
\end{figure}

Figure \ref{fig:observations} (a) and (b) display the token distribution map for H2O and TOVA over a 512-length sequence, both using a cache size of 128. We quantify the values of selected token positions as 1 and 0 otherwise, averaging across the 32 heads to minimize noise. A notable pattern emerges: H2O and TOVA show significant regional biases due to their disregard for the token eviction scope. which may potentially lead to oversimplified interpretations of sequences and impair the models' ability to grasp nuanced interactions, thereby reducing their effectiveness in tasks requiring holistic understanding.

The aforementioned methods aim to reduce the KV cache produced during long text generation, while other studies concentrate on compressing the KV cache of long context prompts during prefilling.
SnapKV \cite{li2024snapkv} retains important tokens based on attention weights alongside their neighboring tokens for additional details. PyramidKV \cite{zhang2024pyramidkv} and PyramidInfer \cite{yang2024pyramidinfer} found that attention is widespread in lower layers and progressively concentrates in higher layers. As a result, they adjust the KV cache size across layers and select tokens in a funnel-like manner. Notably, PyramidKV and PyramidInfer are orthognal to our method.
Although these eviction policies efficiently reduce the size of KV cache, their myopic view of certain regions neglects the comprehensive contextual importance within the broader narrative.

Another line of research is dedicated to structure-guided long context processing. \cite{zhao2022fine} developed a hierarchy for selecting important tokens across layer, \cite{he2024fovea} created a multi-scale tree to efficiently capture the long context dependencies. However, these studies are limited to encoder-based models and cannot be directly applied to pre-trained LLMs without additional tuning, restricting their applicability to generative models.

\section{Preliminary}
In this section, we present our preliminary observations about the input-output dependence in a standard attention layer, which motivates our design of TreeKV. 

Previous work has used Fourier transform to analyze the hidden states of language models \cite{scribano2023dct,he2023fourier}, but those analyses ignore the distributions of the frequency components at different locations of the sequence, due to the lack of locality of the Fourier basis functions. Different from previous work, our analysis uses multi-level discrete wavelet decomposition, which is able to capture the local frequency information at different locations. We first introduce the background knowledge including KV caching and multi-level discrete wavelet decomposition, before presenting our observations.  

\subsection{KV Caching}
\label{kv caching}
Before diving into our method, it's essential to grasp how KV caches are maintained during LLM's inference. The following example illustrates the auto-regressive decoding process in the generation phase.
As KV cache management methods operate consistently across different batches or heads, we have omitted these two dimensions for simplicity. We use $\mathbf{W}_q\in\mathbb R^{d\times d}, \mathbf{W}_k\in\mathbb R^{d\times d}, \mathbf{W}_v\in\mathbb R^{d\times d}$ to denote the weights of the attention modules for a layer, where $d$ is the hidden dimension of the model. 
We use $\mathbf{K}$, $\mathbf{V}$ to denote KV cache and superscript $t$ to denote the generation step.

For a new input $\mathbf{x}^{(t)}\in \mathbb R^{1\times d}$ at generation step $t$, the attention module first transforms it into a set of query, key, and value:
\[
\mathbf{q}^{(t)}=\mathbf{x}^{(t)} \mathbf{W}_Q,
\mathbf{k}^{(t)}=\mathbf{x}^{(t)} \mathbf{W}_K,
\mathbf{v}^{(t)}=\mathbf{x}^{(t)} \mathbf{W}_V.\]

The key and value caches grow linearly by appending the new key and value as:
\[
\mathbf{K}^{(t)}=\begin{pmatrix}
\mathbf{K}^{(t-1)}\\\mathbf{k}^{(t)}
\end{pmatrix}, 
\quad
\mathbf{V}^{(t)}=\begin{pmatrix}
\mathbf{V}^{(t-1)}\\\mathbf{v}^{(t)}
\end{pmatrix}. 
\]

The attention scores $\mathbf{a}^{(t)}$ and the output of the attention $\mathbf{o}^{(t)}$ are computed as follows:
\[
\mathbf{a}^{(t)} = \mathrm{SoftMax}\left(\frac{\mathbf{q}^{(t)} {\mathbf{K}^{(t)}}^\top}{\sqrt d}\right),
\quad
\mathbf{o}^{(t)} = \mathbf{a}^{(t)} \cdot \mathbf{V}^{(t)}.
\]

KV caching reduces redundant computations in LLMs, but as the cache lengthens, the computational cost of attention calculations rises quadratically. This necessitates innovative strategies to optimize the KV cache memory footprint without compromising LLMs performance.

\subsection{Multi-Level Discrete Wavelet Decomposition}
\label{wavelet}
The multi-level discrete wavelet decomposition is a signal processing technique that allows for the representation of a signal at multiple resolution levels using wavelets. Unlike sine and cosine functions used in Fourier transforms, wavelets are localized and can represent both frequency and time (or location in our case) information. In single-level discrete wavelet decomposition, a discrete signal $\mathbf{s}[n]$ is filtered by a pair of low-pass filter $\mathbf{g}[n]$ and high-pass filter $\mathbf{h}[n]$, followed by down-sampling:
\begin{align*}
    & \mathbf{A}_1[n] =  
    \sum_{k=-\infty}^{\infty} \mathbf{s}[k]\mathbf{g}[2n-k]  , \\
    & \mathbf{D}_1[n] = 
    \sum_{k=-\infty}^{\infty} \mathbf{s}[k]\mathbf{h}[2n-k] . 
\end{align*}
The approximation coefficients $\mathbf{A}_1[n]$ represent the low-frequency coefficients of the signal. They capture the main features and general trends of the signal. 
The detail coefficients $\mathbf{D}_1[n]$ correspond to the high-frequency coefficients of the signal, which capture the finer details. We use the Haar wavelet in the following analysis, for which the low-pass filter $\mathbf{g}[n]$ and high-pass filter $\mathbf{h}[n]$ are

\[
\mathbf{g}[i] = \begin{cases}
    \sqrt2/2& i=0,1 \\
    0& \text{otherwise}
\end{cases},
\mathbf{h}[i] = \begin{cases}
    -\sqrt2/2 & i=0\\
    \sqrt2/2 & i=1 \\
    0& \text{otherwise}
\end{cases}.
\]

The single-level reconstruction under the Haar wavelet is 

\[
R(\mathbf{A}_1,\mathbf{D}_1)[n]=\begin{cases}
\frac{\sqrt 2}{2} \left( \mathbf{A}_1\left[\frac{n+1}{2}\right]+\mathbf{D}_1 \left[\frac{n+1}{2}\right] \right) & \text{odd } n \\
\frac{\sqrt 2}{2} \left( \mathbf{A}_1 \left[\frac{n}{2}\right]-\mathbf{D}_1\left[\frac{n}{2}\right] \right) & \text{even } n
\end{cases}.
\]

In multi-level discrete wavelet decomposition, the approximation coefficients are further decomposed through repeated single-level decomposition. Specifically, at level $l>1$, the approximation coefficients $\mathbf{A}_l[n]$ and the detail coefficients $\mathbf{D}_l[n]$ are obtained by filtering and down-sampling $\mathbf{A}_{l-1}[n]$. The coefficients of an $L$-level discrete wavelet decomposition can be organized into a list $[\mathbf{A}_L,\mathbf{D}_L,\mathbf{D}_{L-1},\cdots,\mathbf{D}_1]$, 
where $\mathbf{A}_L$ gives the lowest frequency coefficient of the signal in the decomposition, and $\mathbf{D}_L,\cdots, \mathbf{D}_1$ give the coefficients at progressively higher frequencies. 
Applying the single-level reconstruction from high to low levels recovers the original signal.

\begin{figure}[t]
    \centering
    \includegraphics[width=0.9\linewidth]{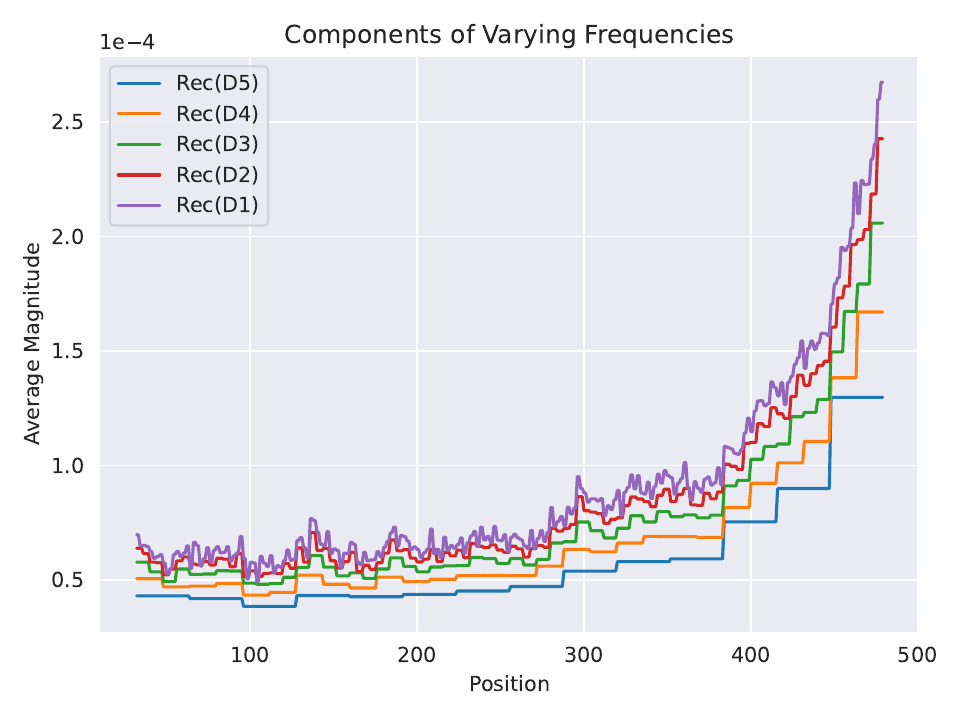}
    \caption{The average magnitude of gathered information's high-frequency components given by wavelet decomposition. We illustrate the top 5 frequency components at the 512th generation step, excluding the first 32 tokens and the last 32 tokens.}
    \label{fig:freq_components}
\end{figure}

\subsection{Observation}
\label{sec: observation}
In full attention, the output at a given position is generated by a weighted sum of the values at all positions up to the given one, where the weights are attention weights. Thus, at the generation step $t$, we view the product of the attention scores and the values up to the position $t$ as signals, which can be formulated as
\[
\mathbf{s} = {\mathbf{a}^{(t)}}^\top \circ {\mathbf{V}^{(t)}}
= \begin{pmatrix}
    \mathbf{a}^{(t)}(1) \mathbf{v}^{(t)}(1) \\
    \vdots \\
    \mathbf{a}^{(t)}(t) \mathbf{v}^{(t)}(t)
\end{pmatrix}.
\]
$\mathbf{a}^{(t)}$ and $\mathbf{V}^{(t)}$ are defined in Section \ref{kv caching}, we analyze it using multi-level discrete wavelet decomposition defined in Section \ref{wavelet}. More precisely, we first apply the multi-level wavelet decomposition along the sequence length dimension to obtain the frequency coefficients at various levels. Next, we reconstruct the signal $\text{Rec}(\mathbf{D}_L)$, which represents frequency components with only the detail coefficients $\mathbf{D}_L$,
\[
\text{Rec}(\mathbf{D}_L) = 
R(R(\cdots R(\mathbf{0},\mathbf{D}_L) \cdots,\mathbf{0}),\mathbf{0}).
\]

Then we average the magnitude of each frequency component across all layers, heads, and samples. Figure \ref{fig:freq_components} shows the average magnitude of each frequency component at different positions for a 5-level wavelet decomposition with the Haar wavelet at the 512th generation step across 2000 samples. 
Since previous work has highlighted the significance of maintaining sinks and recent tokens \cite{xiao2023efficient,han2024lm,zhang2024h2o,liu2024scissorhands,oren2024transformers}, we exclude the first 32 and the last 32 tokens to concentrate on the middle segment of the sequence in the figure. 

We observe that as the positions get closer to the end, all frequency components gradually increase, particularly at higher frequencies. This suggests that the information contributed by a token not only increases, but also tends to diverges more from its neighboring tokens as the position approaches the end of sequence, indicating a smooth transition with increasing complexity and variability from distant to nearby context. This insight inspired our method, resulting in a structure designed to be sparse on the left and dense on the right.

\section{TreeKV}

\begin{figure*}[t]
    \centering
    \includegraphics[width=\textwidth]{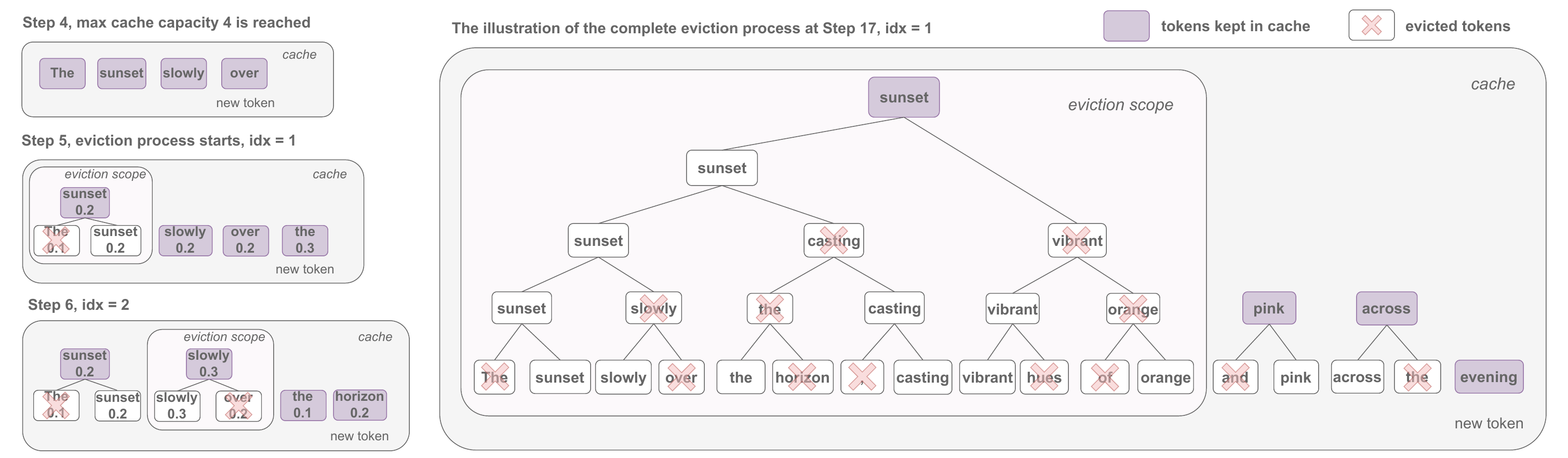}
    \caption{Applying TreeKV cache compression method on a 17-length sequence with cache size 4. Each subplot demonstrates the eviction process and state of the cache at the end of step 4, 5, 6 and 17, respectively, while steps 1-4, showing the cache filling process, are omitted. Variable idx and selection scope are explained in Section \ref{sec: decoding stage}.}
    \label{fig:treekv_generation}
\end{figure*}

This section introduces TreeKV, which organizes keys and values in a tree-like hierarchy to enhance smooth cache compression by leveraging temporal locality. Note that, the structure of TreeKV is inspired by our spectral analysis in Sec \ref{sec: observation}.
Unlike most cache compression methods that focus on either the generation or prefilling phases, TreeKV is applicable to both. We will first outline the compression strategy for the decoding stage, then elaborate its application in the prefilling stage.

\subsection{Decoding Stage}
\label{sec: decoding stage}

\paragraph{Overall Idea} During decoding, the KV pair of new tokens are added sequentially to the cache. Once the cache is at maximum capacity $c$, a tree-based approach strategically evicts KV pair of less important token within a specific eviction scope as generation progresses across layer and head. The parameter $c$ indicates the desired number of key-value pairs to retain in the cache, referred to as cache size. This eviction scope cycles through the cache from distant to nearby context, ensuring a smooth distribution of retained tokens that is sparse on the left and dense on the right. Figure \ref{fig:treekv_generation} provides an illustration of the cache compression process of TreeKV with a maximum cache size set to 4, each subplot demonstrates the eviction process and the state of the cache at the end of step 4, 5, 6 and 17, with steps 1-4 representing the cache filling process. The detailed compression procedure is meticulously outlined in Algorithm \ref{alg:algorithm}.

\paragraph{Importance Score} Each token is associated with its averaged attention weight $\overline{\mathbf{S}}$ (calculated in line 6, 7, 8 and 10 of Algorithm \ref{alg:algorithm}). The averaged attention weight $\overline{\mathbf{S}}$ reflects the importance score of token in the following generation steps, which serves to prioritize which pairs to retain in the cache. We tested various importance criteria in our experiment, including accumulated and normalized attention weights, and found that averaged attention weight delivers the best performance.

\paragraph{Eviction Scope} 
When the maximum cache size $c$ is reached, TreeKV evicts an old KV pair within an ``eviction scope" to maintain the fixed size. The scaler $\mathrm{idx}$ defines this eviction scope, $\mathrm{idx}$ cycles through the cache from distant to nearby contexts, building a tree structure that is sparse on the left and dense on the right. Specifically, the eviction scope is $\left\{\mathrm{idx}, \mathrm{idx+1}\right\}$, indicating that we will remove either the $\mathrm{idx}$-th or the $(\mathrm{idx+1})$-th token in the cache with the lower importance score.
For example, in Figure \ref{fig:treekv_generation}, when the eviction process starts at step 5 with $\mathrm{idx} = 1$, TreeKV evicts the first token in the scope ($1$, $2$) with the lower importance score, resulting in the removal of ``The”. The cache then holds: ``sunset", ``slowly", ``over" and ``the". The scaler $\mathrm{idx}$ cycles through 1, 2, 3, and 4, shifting the eviction scope from left to right to remove older tokens first while prioritizing the recent ones. At step 17, upon the arrival of the token ``evening," the $\mathrm{idx}$ cycles back to $1$, token ``vibrant" from the (``subset", ``vibrant") is evicted. In the figure, we keep the complete eviction process to make the tree structure visible.
Note that our algorithm not just construct trees layer by layer but also facilitates the merging of tokens across different tree levels, maintaining a coherent representation of the input sequence.

\paragraph{Position Encoding} Following StreamingLLM \cite{xiao2023efficient}, the positional encodings are re-assigned after KV eviction. For example, if the current cache contains tokens $\{0, 1, 2, 3, 7, 8, 9$\}, when decoding the 10th token, the assigned positions are $\{0, 1, 2, 3, 4, 5, 6, 7\}$ instead of the original positions $\{0, 1, 2, 3, 7, 8, 9, 10\}$.

\begin{algorithm}[tb]
\caption{Compression by TreeKV}
\label{alg:algorithm}
\textbf{Input}: Inputs $x^{(1)},\cdots,x^{(T)}\in \mathbb R^{1\times d}$, Cache size $c$. \\
\vspace{-1em}
\begin{algorithmic}[1] 
\renewcommand{\algorithmiccomment}[1]{\hfill $\triangleright$ #1}
\STATE Initiate $\mathbf{S},\mathbf{C},\mathbf{K}^{(0)},\mathbf{V}^{(0)}$  as empty tensors, $\mathrm{idx}=1$. 
\FOR{$i$ from $1$ to $T$}
\STATE $\mathbf{q}^{(i)}=\mathbf{x}^{(i)} \mathbf{W}_Q,
\mathbf{k}^{(i)}=\mathbf{x}^{(i)} \mathbf{W}_K, \mathbf{v}^{(i)}=\mathbf{x}^{(i)} \mathbf{W}_V.$
\STATE $\mathbf{K}^{(i)}=
\mathbf{K}^{(i-1)}\cup\{\mathbf{k}^{(i)}\}$.
\STATE $\mathbf{V}^{(i)}=
\mathbf{V}^{(i-1)}\cup\{\mathbf{v}^{(i)}\}$.
\STATE $\mathbf{a}^{(i)} = \mathrm{SoftMax}\left(\frac{\mathbf{q}^{(i)} {\mathbf{K}^{(i)}}^\top}{\sqrt d}\right)$.
\STATE $\mathbf{C} = \left(\mathbf{C}\cup\{0\} \right)+\mathbf{1}$.
\STATE $\mathbf{S} = \left(\mathbf{S}\cup\{0\} \right)+\mathbf{a}^{(i)}$.
\IF{$\text{length}(\mathbf{K}^{(i)})>c$}
\STATE $\overline{\mathbf{S}}=\mathbf{S}/\mathbf{C}$.
\COMMENT{Importance Scores}
\IF {$\overline{\mathbf{S}}_\mathrm{idx}>\overline{\mathbf{S}}_{\mathrm{idx}+1}$}
\STATE  Evict $(\mathrm{idx}+1)$-th elements in $\mathbf{K}^{(i)}$, $ \mathbf{V}^{(i)}$, $\mathbf{C}$,  $\mathbf{S}$.
\ELSE 
\STATE Evict $\mathrm{idx}$-th elements instead.
\ENDIF
\STATE $\mathrm{idx} = (\mathrm{idx}+1)\mod c+1$.
\ENDIF
\ENDFOR
\end{algorithmic}
\end{algorithm}

\paragraph{Conclusion of Decoding Stage} Figure \ref{fig:observations} (c) shows that TreeKV provides a more balanced distributed token map compared to the previous two methods, with a smooth transition from distant coarse-grain to recent fine-grain tokens. Moreover, TreeKV updates the hierarchical structure based on a continuous scoring mechanism, allowing it to adapt to changes in data dynamically. This adaptability ensures that the cache remains relevant to the current context, enabling the system to consistently provide maximum benefit through efficient storage.

\subsection{Prefilling Stage}
\paragraph{Overall Idea} We previously discussed how TreeKV handles eviction during decoding, and this approach also applies to prefilling stage. In this stage, the same principles of importance scoring, eviction scoping and position encoding are employed, but focusing on blocks instead of tokens. To enable the model to operate efficiently, all blocks are processed simultaneously, with the importance scores and eviction scopes computed concurrently. This pre-computation allows the model to swiftly select the most important block within each eviction scope.

\paragraph{Block-level Eviction}
Most context compression techniques rely on token-level selection \cite{zhang2024h2o,wang2024xl3m}, overlooking the fact that either critical or irrelevant information is often spatially clustered \cite{li2024snapkv}. Selecting individual tokens can compromise contextual integrity and computational speed. Therefore, we adopt a block-level eviction policy for prompt compression tasks, applying our algorithm on blocks instead of tokens. TreeKV first divides the prompt into multiple blocks of size $b$, with each block representing one token in Fig \ref{fig:treekv_generation}. Following \cite{li2024snapkv}, we use the last block of the prompt as the observation window to \textit{query} the entire sequence. We obtain the importance score of each block by calculating the averaged importance scores for the $b$ tokens within that block.

\paragraph{Conclusion of Prefilling Stage} In summary, the prefilling stage of TreeKV mirrors the strategies employed during the decoding stage by focusing on blocks rather than tokens. Simultaneous block computing allows TreeKV to efficiently managing KV pairs while maintaining integrity and contextual relevance.

\section{Experiments}
Our evaluation of TreeKV demonstrates its superiority over existing methods in both prefilling and decoding phases. We first assess its performance on the language modeling task with PG19 \cite{rae2019compressive} and OpenWebText2 \cite{gao2020pile} for long text decoding. Additionally, we demonstrate that TreeKV can reliably handle texts exceeding 10 million tokens, achieving the lowest perplexity among all baseline models.
Finally, we assess TreeKV's prompt compression ability during prefilling using LlaMa-3.2-1B-Instruct on the Longbench \cite{bai2023longbench} benchmark.

\subsection{Setup}
For language modeling task, we use Llama-2-7B \cite{touvron2023llama} pre-trained with 4K context length as base model considering its popularity and outstanding performance. We evaluate perplexity using a sliding window approach with a stride of 2048 for PG19 and 1024 for OpenWebText2 respectively. For language understanding tasks on Longbench, we truncate the inputs to 32k in the same manner as SnapKV \cite{li2024snapkv}. We employ Llama-3.2-1B-Instruct as our base models. All the experiments utilize bf16 precision on Nvidia RTX4090 GPUs.

\begin{table}[t]
\centering
\begin{tabular}{@{}lccc@{}}
\toprule
                & \multicolumn{3}{c}{\textbf{PG19}} \\ \cmidrule(lr){2-4} 
Context Length  & 4k        & 8k        & 16k          \\
Budget Ratio & 25.0\%        & 12.5\%        & 6.3\%       \\ 
\midrule
Full Attn       &  6.84      &    $>10^3$       &   OOM         \\ \midrule
StreamingLLM    & 7.19      &  7.19     & 7.19         \\
H2O             & 7.06      &   7.08        & 7.25         \\
TOVA            &  \textbf{7.00}     & 7.06       & 7.15         \\
TreeKV (ours)   & 7.02      &  \textbf{6.88}     & \textbf{6.91}          \\ \bottomrule
\end{tabular}
\caption{Sliding window perplexity of different context window extension methods using Llama-2-7B model on PG19. The cache size of all the efficient methods is set to 1024.}
\label{tab: pg19}
\end{table}

\begin{table}[t]
\centering
\begin{tabular}{@{}lccc@{}}
\toprule
                & \multicolumn{3}{c}{\textbf{OpenWebText2}} \\ \cmidrule(lr){2-4} 
Context Length  & 4k        & 8k        & 16k         \\
Budget Ratio & 25.0\%        & 12.5\%        & 6.3\%      \\ 
\midrule
Full Attn       &  5.44         &    $>10^3$       &   OOM           \\ \midrule
StreamingLLM        &  5.78           &   5.62          &  5.31        \\
H2O               &  \textbf{5.60}           &    5.48         &  5.25       \\
TOVA             &    5.62         &    5.50         &  5.24        \\
TreeKV (ours)      &  \textbf{5.60}           &   \textbf{5.45}          &  \textbf{5.18}        \\ \bottomrule
\end{tabular}
\caption{Sliding window perplexity of different context window extension methods using Llama-2-7B model on OpenWebText2. The cache size is set to 1024 for all the experiments.}
\label{tab: owt2}
\end{table}

\begin{figure}[t]
    \centering
    \includegraphics[width=\linewidth]{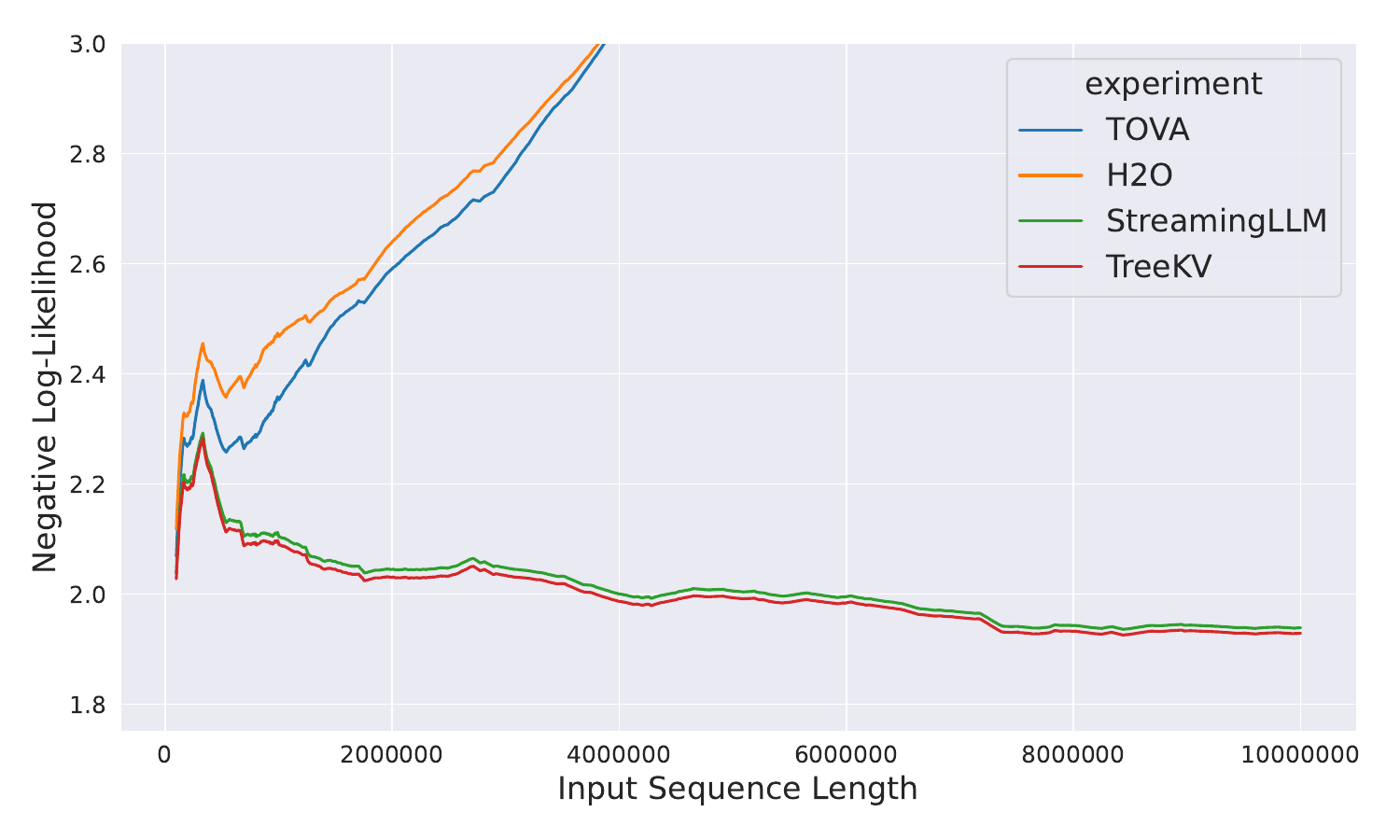}
    \caption{We create a 10M length sequence by concatenating the first 50 books from the PG19 test set and applying 4 efficient decoding methods: TOVA, H2O, StreamingLLM, and TreeKV on it. We then plot the negative log-likelihoods against sequence length.}
    \label{fig:5m}
\end{figure}

\begin{table*}[t]
\centering

\resizebox{\textwidth}{!}{
\begin{tabular}
{@{}l
c@{\hspace{0.05ex}}c@{\hspace{0.05ex}}c
c@{\hspace{0.05ex}}c@{\hspace{0.05ex}}c
c@{\hspace{0.05ex}}c@{\hspace{0.05ex}}c
c@{\hspace{0.05ex}}c@{\hspace{0.05ex}}c
c@{\hspace{0.05ex}}c
c@{\hspace{1ex}}c
c@{}}

\specialrule{1pt}{0pt}{2pt}
\multirow{3}{*}{Method} & \multicolumn{3}{c}{Single-Document QA} & \multicolumn{3}{c}{Multi-Document QA}& \multicolumn{3}{c}{Summarization}& \multicolumn{3}{c}{Few-shot Learning}& \multicolumn{2}{c}{Synthetic} & \multicolumn{2}{c}{Code} & \multirow{3}{*}{Avg.} \\

\cmidrule(lr){2-4}\cmidrule(lr){5-7}\cmidrule(lr){8-10}\cmidrule(lr){11-13}\cmidrule(lr){14-15}\cmidrule(lr){16-17}

& \rotatebox[origin=c]{30}{NrtvQA} & \rotatebox[origin=c]{30}{Qasper} & \rotatebox[origin=c]{30}{MF-en} & \rotatebox[origin=c]{30}{HotpotQA} & \rotatebox[origin=c]{30}{2WikiMQA} & \rotatebox[origin=c]{30}{Musique} & \rotatebox[origin=c]{30}{GovReport} & \rotatebox[origin=c]{30}{QMSum} & \rotatebox[origin=c]{30}{MultiNews} & \rotatebox[origin=c]{30}{TREC} & \rotatebox[origin=c]{30}{TriviaQA} & \rotatebox[origin=c]{30}{SAMSum} & \rotatebox[origin=c]{30}{PCount} & \rotatebox[origin=c]{30}{PRe} & \rotatebox[origin=c]{30}{Lcc} & \rotatebox[origin=c]{30}{RB-P} & \\

\arrayrulecolor{black}\midrule
\multicolumn{18}{c}{\textcolor{gray}{Cache Size = Full}} \\
\arrayrulecolor{black!20}\midrule
\multirow{1}{*}{\textcolor{gray}{FullKV}}
& \textcolor{gray}{20.78} & \textcolor{gray}{24.05} & \textcolor{gray}{42.00}    
& \textcolor{gray}{36.74} & \textcolor{gray}{29.30} & \textcolor{gray}{21.29}     
& \textcolor{gray}{28.77} & \textcolor{gray}{22.06} & \textcolor{gray}{25.41}  
& \textcolor{gray}{64.00} & \textcolor{gray}{80.05} & \textcolor{gray}{39.58}   
& \textcolor{gray}{4.50} & \textcolor{gray}{4.09}  
& \textcolor{gray}{39.11} & \textcolor{gray}{43.21}  
& \textcolor{gray}{32.86} \\

\arrayrulecolor{black}\midrule
\multicolumn{18}{c}{Cache Size = 2048} \\
\arrayrulecolor{black!20}\midrule

\multirow{1}{*}{H2O}   
& 19.34  & 21.59 & 40.81    
& 35.43  & 27.57 & 18.71     
& 23.18  & 21.33 & 24.92  
& 45.50  & 80.12 & 37.69 
& 3.50 & 4.09
& 37.54 & 42.28
& 30.23 \\

\arrayrulecolor{black!20}\midrule

\multirow{1}{*}{SnapKV} 
& \textbf{20.82}  & 22.23 & \textbf{42.34}    
& \textbf{38.12}  & \textbf{29.17} & \textbf{21.16}     
& \textbf{23.96}  & \textbf{22.15} & \textbf{24.97}  
& \textbf{63.00}  & 80.90 & \textbf{38.40} 
& 3.50 & \textbf{4.18}
& 30.60 & 38.47
& 31.50 \\

\arrayrulecolor{black!20}\midrule

\multirow{1}{*}{TreeKV}  
& 20.40  & \textbf{23.20} & 41.89    
& 35.62  & 27.49 & 18.93     
& 23.34  & 21.19 & 24.70  
& 61.00  & \textbf{81.00} & 37.75 
&\textbf{4.50} & 3.82
& \textbf{39.11} & \textbf{43.28}
& \textbf{31.70} \\

\arrayrulecolor{black}\midrule

\multicolumn{18}{c}{Cache Size = 8192} \\
\arrayrulecolor{black!20}\midrule

\multirow{1}{*}{H2O}   
& 21.19  & 23.85 & 43.19    
& 37.26  & 29.22 & 20.60     
& 27.61  & \textbf{22.18} & \textbf{25.53}  
& 63.50  & 80.90 & \textbf{39.68}
& \textbf{3.50} & \textbf{4.50}
& 38.60 & 43.29
& 32.79 \\

\arrayrulecolor{black!20}\midrule

\multirow{1}{*}{SnapKV} 
& 21.26  & 24.03 & \textbf{43.40} 
& 37.22  & \textbf{28.82} & \textbf{22.20} 
& \textbf{28.18}  & 22.09 & \textbf{25.53}  
& \textbf{64.00}  & 80.85 & 38.57 
& \textbf{3.50} & \textbf{4.50}
& 30.69 & 37.72
& 32.04 \\

\arrayrulecolor{black!20}\midrule

\multirow{1}{*}{TreeKV} 
& \textbf{21.48}  & \textbf{24.47} & 42.19    
& \textbf{37.62}  & 28.75 & 21.69     
& 27.76  & 21.64 & \textbf{25.53}
& \textbf{64.00}  & \textbf{81.01} & 39.13 
& \textbf{3.50} & 4.09
& \textbf{38.65} & \textbf{43.29}
& \textbf{32.80}\\

\arrayrulecolor{black}\bottomrule

\end{tabular}
}
\caption{Performance comparison of H2O, SnapKV and TreeKV on Longbench using LlaMa-3.2-1B as base models. Results are reported with cache sizes of 2048 and 8192. SnapKV results were obtained using their officially released code, while H2O was implemented by the authors. The best scores are highlighted in bold. Our model outperforms H2O and SnapKV on all the cache sizes. The results of Full attention models are shown at the top of the table.}
\label{tab: longbench}
\end{table*}

\subsection{Language Modeling}
The language modeling task assesses LLMs' ability to predict the next token from the preceding context. We report perplexity on two datasets: PG19 test set \cite{rae2019compressive} and OpenWebText2. These two datasets are both commonly used datasets for evaluating long-range language models. The PG19 test set consists of 100 full-length books, each averaging 113k tokens. The OpenWebText2 dataset is derived from the Pile dataset \cite{gao2020pile}, from which we randomly selected 100 samples from the test set, averaging 18k tokens each, for evaluation. 

We compare TreeKV with 4 baseline methods: efficient decoding policies like StreamingLLM \cite{xiao2023efficient}, H2O \cite{zhang2024h2o}, and TOVA \cite{oren2024transformers}, as well as a full attention method that caches all keys and values. We assessed these five methods using sequences of context lengths 4k, 8k, and 16k, while keeping a cache size of only 1k. The 1k cache kept from each method consists of three components: 4 initial tokens, 508 recent tokens, and 512 method-selected tokens.

TreeKV surpasses all baselines when context length exceeds the pre-trained limit of the LLM and also performs competitively within that limit. The results are shown in Table \ref{tab: pg19} and \ref{tab: owt2}. Although, TreeKV is 0.02 behind TOVA for 4k-length contexts in PG19 but soon turn the tide with longer contexts. On OpenWebText2, we achieve the best perplexity across all context lengths. 
Our method shows the most significant improvement across all the longest length, surpassing the second-best TOVA by $3.6\%$ and $1.1\%$ on the PG19 and OpenWebText2 datasets respectively, with 16k context length and up to 16x reduction in KV cache.

Following \cite{xiao2023efficient,han2024lm,zhang2024h2o}, we also thoroughly examine TreeKV to test if a LLM pre-trained with a 4k context window can effectively perform language modeling task on exceptionally extended text. 
We concatenate all the 100 books in PG19 test set to create a 10M length example and evaluate its generation capability using Llama-2-7B with StreamingLLM, H2O, TOVA, and TreeKV. Figure \ref{fig:5m} shows the NLL curves for input lengths from 0.1M to 10M. The curves reveal that TOVA and H2O suffer performance degradation with longer sequences, leading to significantly worse NLL than StreamingLLM and TreeKV. In contrast, TreeKV consistently outperforms all other baseline methods including StreamingLLM, demonstrating superior capability with longer inputs.

\subsection{Long Context Understanding}
Unlike most KV cache compression methods that focus on either the generation or prefill stages, TreeKV is applicable to both. We conduct experiments on long context understanding tasks using the Longbench \cite{bai2023longbench} benchmark. Longbench is a multi-task benchmark that includes a wide range of long-context tasks to assess model capabilities in handling extended textual input. Longbench consists of 16 tasks across 6 categories: single-document QA, multi-document QA, summarization, few-shot learning, synthetic tasks, and code completion. The average length across all the sub-datasets is around 11k.

We compare our method against three baseline approaches: H2O \cite{zhang2024h2o}, a method that also suitable for both prefilling and generation stages; SnapKV \cite{li2024snapkv}, a state-of-the-art method for long context understanding; and a full attention method that caches all keys and values.

Table \ref{tab: longbench} summarizes the performance metrics for all tasks and cache sizes in the Longbench benchmark. Overall, our model achieved an average improvement of 0.74 over H2O and 0.48 over SnapKV, demonstrating its superior capacity to retain and recall relevant information from extended text. TreeKV underperforms SnapKV in Multi-Doc QA and Summarization tasks at cache size 2k. Since TreeKV algorithmically prioritizes recent tokens when the cache size is limited, we hypothesize that limiting the tree height to balance early and recent token retention could mitigate this issue, which will explore in future. TreeKV did not surpass the full attention model, which leaves space for us to improve on with minimal resource requirements.

\subsection{Ablation Study}

An important question is: what is the key component of our approach — the tree structure or the attention weight-based token selection mechanism? To investigate this, we modified our method to consistently select the leftmost token within the eviction scope instead of the one with the highest attention weight. This change allowed us to isolate the effect of the tree structure itself, facilitating a focused examination of how token hierarchy affects model performance, independent of the variability introduced by weight differentiation.

\begin{figure}[t]
    \centering
    \includegraphics[width=\linewidth]{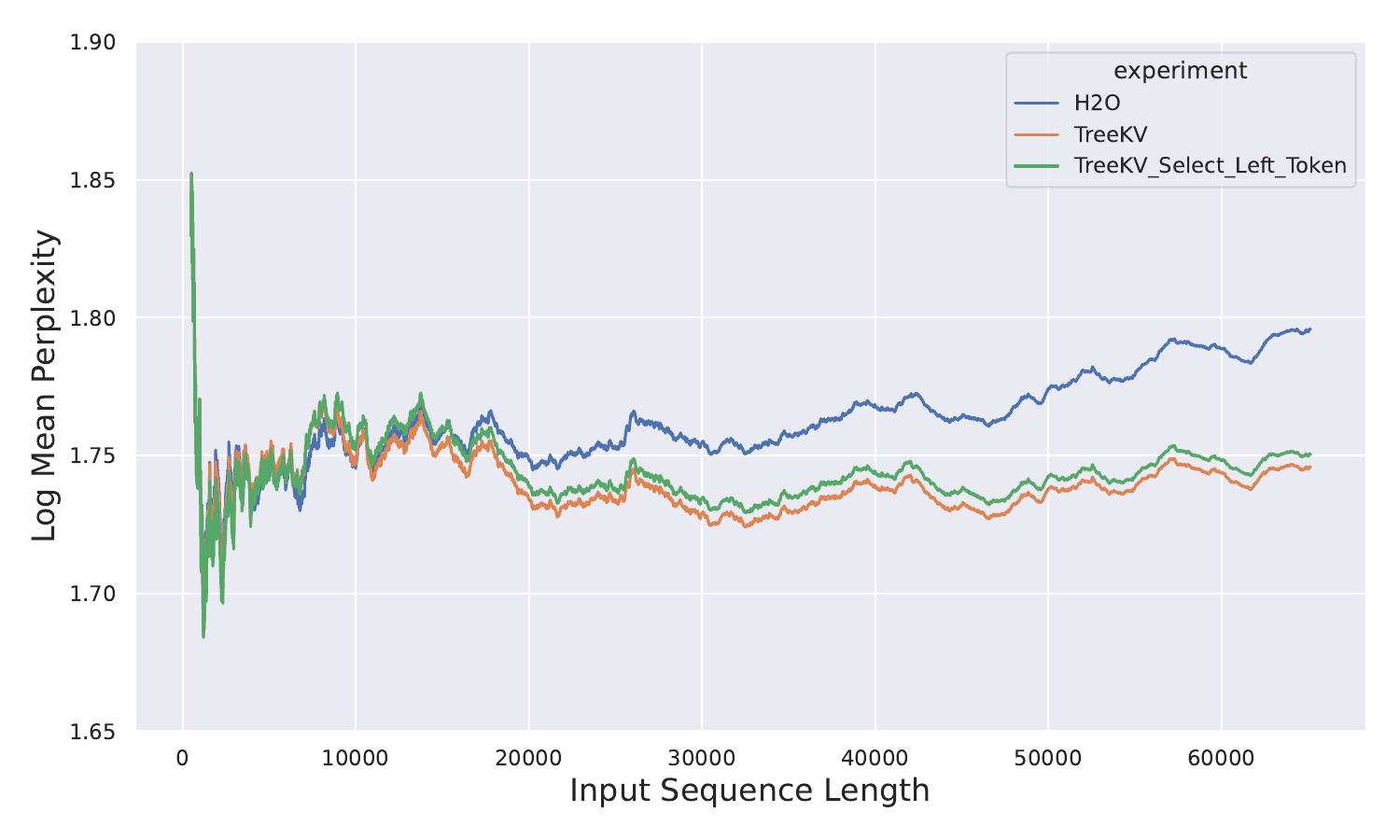}
    \caption{We plot the log mean perplexity of the first book from PG19, which has a length of 65k. In the TreeKV\_Select\_Left\_Token experiment, we consistently select the left token within the eviction scope instead of the one with the highest attention weight.}
    \label{fig:ablation}
\end{figure}

In Figure \ref{fig:ablation}, we present the log mean perplexity of the first book from PG19, which is 65k tokens long. We compare three methods: 1) H2O, which greedily selects tokens based on their cumulative attention weights. 2) TreeKV, which employs the average attention weights to guide evictions through tree structure; and 3) TreeKV\_Select\_Left\_Token, a variant of TreeKV that prioritizes the leftmost tokens without using attention scores. The results show that the consistent token selection strategy results in minor variations in perplexity, indicating the tree structure's significant role in shaping the model's decision-making. In conclusion, our investigation confirms that the tree structure is a crucial component of our approach.

\section{Conclusion}
In this paper, we first explore the frequency representations of information collected during generation using multi-level wavelet decomposition, leading to the development of TreeKV, a training-free method that utilizes a tree structure for smooth cache compression. Our evaluation shows that TreeKV outperforms existing methods in both prefilling and generation setups.

\newpage

%% The file named.bst is a bibliography style file for BibTeX 0.99c
\bibliographystyle{named}
\bibliography{ijcai25}

\end{document}